\documentclass[a4paper]{article}
\usepackage{anyfontsize}
\usepackage{INTERSPEECH2019}
\usepackage[table]{xcolor}
\usepackage{booktabs}
\usepackage{graphicx}
\usepackage[noadjust]{cite}
    
\usepackage{hyperref}
\usepackage{caption}

\usepackage{float}
\title{  End-to-End Neural Transformer Based Spoken Language Understanding }
\name{Martin Radfar, Athanasios Mouchtaris,  and Siegfried Kunzmann}
\address{Alexa Machine Learning, Amazon }
\email{\{radfarmr,mouchta,kunzman\}@amazon.com}

\begin{document}

\maketitle
\begin{abstract}

Spoken language understanding (SLU) refers to the process of inferring the semantic  information from audio signals.  While the neural transformers consistently deliver the best performance among the state-of-the-art neural architectures in field of natural language processing (NLP),  their merits in a closely related field,\emph{ i.e.},  spoken language understanding (SLU)  have not beed investigated. In this paper,
we introduce an end-to-end neural transformer-based SLU model that can predict the variable-length domain, intent, and slots  vectors embedded in an audio signal with no intermediate token prediction architecture. This new architecture leverages the self-attention mechanism by which the audio signal is transformed to various sub-subspaces allowing to extract the semantic context implied by an utterance. Our end-to-end  transformer SLU predicts the domains, intents  and slots in the Fluent Speech  Commands dataset with accuracy equal to 98.1 \%, 99.6 \%, and 99.6 \%, respectively and outperforms the SLU models that leverage a combination of recurrent and convolutional neural networks by 1.4 \%   while the size of our model is  25\% smaller than  that of these architectures.  Additionally, due to independent sub-space projections in the self-attention layer,  the model is highly parallelizable which makes it a good candidate for on-device SLU. 

\end{abstract}

\section{Introduction}
Spoken language understanding (SLU) systems  extract   semantic information from a spoken utterance  by machine \cite{wang2005spoken}.  The
 Air Travel Information System (ATIS)  was the first  SLU model built based on a cascade of a speech recognizer, a language model,  a semantic extractor-SQL generator  (NLU) in 1990 \cite{price1990evaluation}.    Thirty years after ATIS, designing an end-to-end  (E2E)  neural SLU  that can replace  the ASR+NLU-based SLU technology still remains a challenge \cite{lee2015spoken,lugosch2019speech,haghani2018audio,renkens2018capsule,chen2018spoken,tomashenko2019recent,mesnil2013investigation,mesnil2014using,yao2014spoken,vu2016bi,liu2017topic,chen2019transfer,huang2019adapting,tomashenko2020dialogue,dinarelli2020data,wang2020large}. Ideally, we would like to have an all-neural model whose layers project the audio signal to hidden semantic representations, the-so-called "thought vectors" \cite{jeffhinton1} to infer the domain, intent, and slots implied by the audio signal. 
 
 To achieve this goal, several groups conducted experiments using non-ASR awareness  E2E SLU\cite{qian2017exploring,chen2018spoken,serdyuk2018towards,lee2015spoken,liu2017topic}.  These models usually apply multiple stack of RNNs  \cite{qian2017exploring,lugosch2019speech,mesnil2014using,yao2014spoken,vu2016bi,serdyuk2018towards} to encode the entire utterance to a vector which is fed to a fully connected feedforward neural network followed by a soft-mask or a max-pool layer to identify the domain, intent, or slot.  These models treat each unique combination of domain, intent, and slots as an output label. For this reason, we call this type of E2E SLU  classification-based approaches. The limitation of classification-based approaches is that  the combination of domains, intents, and slots may grow exponentially, subsequently we deal with a classification problem with many number of output labels; moreover, the number of intents is not usually fixed which makes usability of classification-based approaches more limited. 
 
 A natural approach to deal with the variable-length output for E2E SLU is to use the sequence-to-sequence (seq2seq) neural models \cite{NIPS2014_5346}.  In \cite{haghani2018audio}, several seq2seq  architectures are proposed for E2E SLU, among which the authors found the model that incorporates  an ASR-awareness module in form of  a multi-task learner delivers the best performance. The finding is further supported by a recent proposed pre-trained ASR-awareness architecture \cite{lugosch2019speech}.

In  ASR, the input and output sequences are ordered and monotonic.  As such, we don't need the entire utterance to decode the transcription. In contrast, when extracting semantic information from audio signals, we usually need to scan the entire audio. Similar to neural machine translation \cite{bahdanau2014neural, devlin2018bert}, the E2E  SLU models can massively benefit from the attention mechanism.  In attention mechanism, the  encoder generates the outputs by incorporating the hidden representations from all time steps and hence allows the output to pay attention to inputs at all time steps \cite{vaswani2017attention}.  The neural attention models  have demonstrated promising results in ASR as well \cite{chan2016listen,dong2018speech}.

 The transformers are seq2seq, non-recurrent, self-attention neural models that have been used in neural machine translation  as well as NLU with great success \cite{vaswani2017attention,bahdanau2014neural, devlin2018bert,vaswani2018tensor2tensor,chen2019transfer}. In this paper, we leverage the transformer architecture for E2E SLU. Neural transformers have several distinct features which make them suitable candidate  for SLU task: (a) The transformers use the self-attention mechanism that allows to compute the correlation in each sublayer between all pairs of time steps both in  the encoder and decoder. (b) Sub-spaces projection by self-attention helps extract semantic context from audio frames. (c) The transformers can benefit from distributed training because linear transformation in self-attention can be parallelized.  (d) Compared to RNN  models \cite{lugosch2019speech,mesnil2014using,yao2014spoken,vu2016bi}, the transformers have less number of parameters.  Our model works both in a classification-based mode and in a hierarchical mode that allows to  decode variable length domain, intent and slot vectors.  We compare this new architecture with  E2E SLU models that use both RNN and CNN on the recently publicly released dataset called Fluent Speech  Commands \cite{lugosch2019speech,serdyuk2018towards}. Our results show that the transformer-based SLU outperforms the RNN+CNN based model, while it has  less numbers of parameters.
 
  The rest of this paper is organized as follows:  Section 2 formulates the problem.  In Section 3,  we describe the transformer based SLU in details. Section 4 gives the details of experiments and results. Finally, we draw the conclusion and give the future directions in Section 5.
 
\section{Problem Formulation}
\begin{figure}[t]
  \centering
  \includegraphics[width=.7\linewidth]{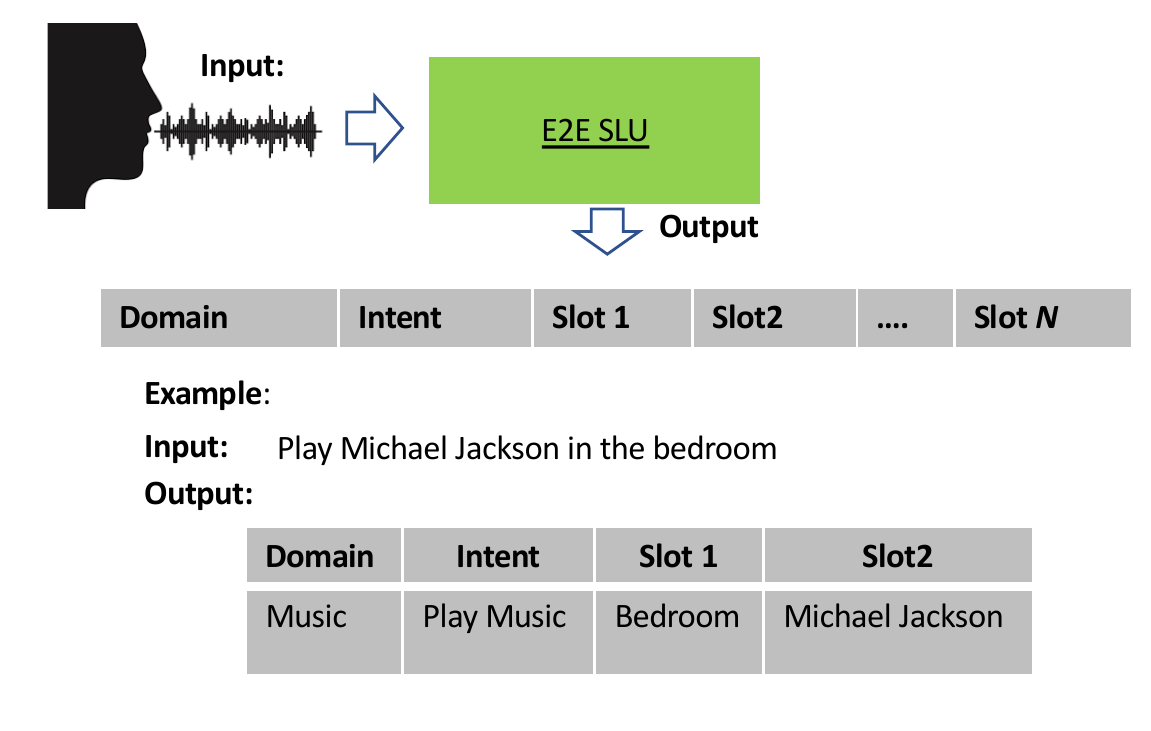}
  \caption{ The SLU task: infer the domain, intent, and slots given an  utterance}
  \label{fig:slutask}
\end{figure}

 Given an utterance transformed to the feature space $X_{p\times T}$, where $p$ and $T$ are the dimension of  the feature space and the number of frames, respectively, we would like to design an all-neural architecture that predicts the output vector $y_{(M+2)\times 1}=[y_D, y_I, y_{s_1}, y_{s_2}, \ldots, y_{s_M} ]^T$ which consists of the domain, intent, and slots  implied by this utterance.  Generally, for a given utterance, we have one domain, one intent, and multiple slots; an example of  a typical SLU task depicted in Figure~\ref{fig:slutask}. The total number of output labels for this classification problem  could be as high as  $N_D \times N_I \times N_{s_1} \times N_{s_2}\times \ldots \times N_{s_M} $ where $N_D, N_I, N_{s_i}$ denote the number of domains,  the number of intents, and the number of  the $i$th slot. Hence, given $N$  input-output training samples  ($X^j_{p\times T,}y^j_{(M+2)\times 1}), j =1, \ldots, N$, we would like to train a neural model that can predict the domain, intent, and slot for a test utterance. 
 
 In this paper,  we consider two models: 1- A classification-based model in which we have $N_D \times N_I \times N_{s_1} \times N_{s_2}\times \ldots \times N_{s_M} $ classes and we assign one output label for each audio file.  In other words, each unique combination of domain, intent, and slots is considered as an output class. In this model, the decoder is basically a fully connected feedforward neural network followed by a softmax. This model works well when the number of classes are limited. 2- A hierarchical model  is similar to seq2seq neural architectures \cite{NIPS2014_5346} and  works as follows:  Let the vector 
 $y_{(M+2)\times 1}=[y_D, y_I, y_{s_1}, y_{s_2}, \ldots, y_{s_M} ]^T$  denote the domain, intent, and slots implied by  $X_{p\times T}$.  We augment $y_{(M+2)\times 1}$ by two  symbols which represent the start and end of this phrase ($sop$ and $eop$), i.e.   $\hat{y}_{(M+2)\times 1}=[sop,y_D, y_I, y_{s_1}, y_{s_2}, \ldots, y_{s_M}, eop ]^T$. In the decoding step, we first predict  $y_D$ by inputing $sop$ to the input of the decoder along with the encoder output. Next, the predicted $y_D$ is fed to the decoder along with the encoder output to predict $ y_I$ and this process is repeated until the predicted output is $eop$.   We call this model E2E hierarchical transformer SLU to distinguish from the  E2E classification-based transformer SLU.

\section{Model}
The proposed SLU model is based on the neural transformers. The transformer architecture consists of three sub-modules: embedding, encoder, and decoder. The embedding is a  linear transformation on the input vector followed by adding a positional encoding vector to the transformed input to properly encode the order of the frames. The encoder consists of two blocks: self-attention, and a fully connected feed-forward neural network (FFNN).  The stack of self-attention and FFNN comprises a layer. Depending on application, we use different number of layers.   The self-attention carries out multiple sub-space transformations; more specifically, the self-attention sub-layer computes the correlation between all frame pairs in  two  sub-spaces and outputs the sum of weighted sub-spaces vectors .  The weight values determine the amount  of correlation between a frame and others. In order to capture more hidden correlation representation, this process  is repeated  in parallel in different subspaces (the so-called multi-head attention) and  resulting attention vectors are appended and linearly transformed  and are fed to the FFNN. In oder to speed up the learning process and prevent the network  forgetting the previous layer representations, we apply layer normalization \cite{ba2016layer} and residual connection \cite{he2016deep} at the outputs of the  self-attention, and FFNN sub-layers.
 
 The above description can be represented based on a linear algebraic view which mostly consists of matrix sub-space analysis and transformation. More specifically, let $\times$ and $\odot$ denote the matrix multiplication and dot product, respectively. 
Also, let $ x^i  \in R^{ p },  i = 1,2, \ldots, T $ denote the $i $th spectral feature  vector of an utterance which is segmented to $T$  overlapping frames and transformed to the spectral features (here we use low frame rate log  STFT).  

The first step is embedding in which the input vector  $ x^i_{p \times 1}$ is  projected to a smaller space ($d <p$) using the matrix $W^e_{d \times p}$ that yields 
 
 \begin{equation}
  y_{d \times 1} = W^e_{d \times p}\times x_{p \times 1} + p_{d \times 1}
\end{equation}
where $p_{d \times 1}$ is a positional encoding vector. We note that all matrices denoted by $W$ are learned during the training using the backpropagation algorithm. 

Next, the new embedding vector is projected to three subspaces using  $W^q_{n \times d} $, $W^k_{n \times d}$, and $W^v_{n \times d}$ matrices which metaphorically named query, key, and value matrices:
 \begin{equation}
 \begin{aligned}
 q^i_{n \times 1} = W^q_{n \times d}  \times y^i_{d \times 1} \\
k^i_{n \times 1} = W^k_{n \times d}  \times  y^i_{d \times 1} \\
v^i_{n \times 1} = W^v_{n \times d} \times  y^i_{d \times 1} \\
\end{aligned}
 \end{equation}
For the $i$th frame, we compute the weighted sum of $v^j$ , $j=1,\ldots,T$, where weights are obtained by dot product and $q^i$ and $k^j$ followed by the soft-max operation and a division to the square root of $n$. Speaking mathematically, this can be written as
 \begin{equation}
z^i_{d \times 1}= W ^c _{d\times n} \times \Big( \displaystyle \sum_{j =1} ^T   \frac{\text{Softmax}( {q^i}_{n \times 1} \odot  k^j_{n \times 1} )}{\sqrt{n}} v^j_{n\times 1} \Big)
 \end{equation}
 where $W ^c _{d\times n} $ projects the resulting vector to a $d$-dimensional space.
 Next, the layer normalization and residual connection are applied to the output of the self-attention sublayer  given by 
  \begin{equation}
 h^i_{d \times 1} = \text{Norm}(z^i_{d \times 1}+ y^i_{d \times 1}) 
 \end{equation}
 and lastly, a two-layer FFNN  transforms $h^i_{d \times 1} $ to
 \begin{equation}
 s^i_{d \times 1} = \text{max}(0, h^i_{d \times 1}W^{f_1} )\times W^{f_2} 
 \end{equation}
followed by another layer normalization and residual connection:
 \begin{equation}
 r^i_{d \times 1} = \text{Norm}(s^i_{d \times 1}+ h^i_{d \times 1}) .
  \end{equation}
  We use a stack of several layers in our transformer; in this case, $y_{d \times 1} \leftarrow  r^i_{d \times 1} $ and the above process is repeated for the next encoder until we reach to the last stack in the encoder.  We also use multi-head attention in which  the process of transforming  $y_{d \times 1}$  to sub-spaces is carried out in parallel with introducing $L$ tuples of ( $W^{q,l}_{n \times d} $, $W^{k,l}_{n \times d}$, and $W^{v,l}_{n \times d}$), for $l= 1,\ldots, L$ where $L$ denotes the number of heads. The outputs of these heads, $z^{i,l}_{d \times 1}$ for $l= 1,\ldots, L$ are appended $[z^{i,1}_{d \times 1},\ldots, z^{i,L}_{d \times 1}]$ and transformed to a $d$-dimensional vector using  the matrix $W_{d\times (L d)}$.  This operation outputs the $d$-dimensional output vector $z^i_{d \times 1}$ which will be fed to FFNN.

 The decoder of the transformer has a similar structure to the encoder with two major differences; first, the decoder has an encoder-decoder attention sublayer in addition to the self-attention layer which is placed between the self-attention sub-layer and FFNN.  The encoder-decoder attention receives  $r^i_{d \times 1}$ from the encoder and $ h^i_{d \times 1} $ from the self-attention sub-layer of  the decoder and transforms them similar to the self-attention sub-layer but with these inputs: 
  \begin{equation}
 \begin{aligned}
 q^i_{n \times 1} = W^q_{n \times d}  \times h^i_{d \times 1} \\
k^i_{n \times 1} = W^k_{n \times d}  \times  r^i_{d \times 1} \\
v^i_{n \times 1} = W^v_{n \times d} \times  r^i_{d \times 1} .\\
\end{aligned}
 \end{equation}
The second difference between the encoder and the decoder is  the masking process. Because the decoder does not have  information from the future decoded sequences, it will mask them during the decoding process. The rest of process in the decoder is the same as in the encoder.  Finally, the output of the decoder is fed to a fully connected neural network  followed by a soft mask to generate the posterior probabilities for domain, intent, and slots.
 \begin{figure}[h]
  \centering
  \includegraphics[width=.7\linewidth]{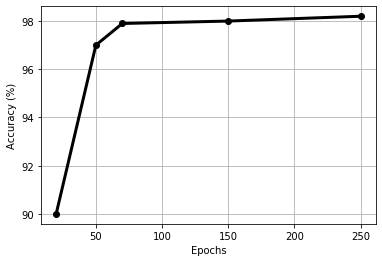}
  \caption{The transformer accuracy v.s. number of epoch for the domain class}
  \label{fig:accuracyvsepoch}
\end{figure}
 
  \section{Experiments}
 \subsection{Feature generation}
 We used 80-dimensional log short time Fourier transform vectors obtained by segmenting the utterances with a Hamming window of the length 25 ms and frame rate of 10 ms. The four frames are stacked with a skip rate of 3, resulting  320-dimensional input features. We used compute-fbank-feats in Kaldi \cite{Povey_ASRU2011} to generate the features; these features are normalized using the global mean and variance normalization.
 \subsection{Transformer parameters}
 The numbers of layers for the encoder and the decoder were set to five and one, respectively; we used three heads in each attention layer and sub-space dimension, i.e. $n$, was set to 64. The dimension of the model , i.e. $ d$ , and the inner dimension of  FFNN were set to 128 and 512, respectively. The dropout rate and label smoothing were set to 0.1.  We used  Adam optimizer \cite{kingma2014adam} with $\beta_1$ = 0.9, $\beta_2$= 0.98, and $\epsilon$ = 1e-9.  We found that the transformers' performance  is, in general, very sensitive to learning rate.
We used the learning rate  defined in \cite{dong2018speech} in which we increased the learning rate  linearly to a threshold and then it  was reduced in a non-linear fashion per step.
We obtained our best results by setting pre-defined factor $k$ and warm-up rate $w$ to 0.95 and 18,000, respectively. We trained our model with different number of epochs ranging from 50 to 1,500.
After a certain number of steps, the model is evaluated on  the evaluation data and the best model is saved.  We found that the loss decreases very slowly after 50 epochs  but the results are consistently improved up to 250 epochs after which the improvement is very marginal (see, e.g. Figure~\ref{fig:accuracyvsepoch}). 
 Our implementation is adapted from a Pytorch implementation which originally designed for speech recognition \footnote{https://github.com/kaituoxu/Speech-Transformer} .


 \subsection{Data set}
We used the publicly available Fluent Speech  Commands (FSC) dataset \cite{lugosch2019speech} to train and evaluate our model and compare with  models tested on the same dataset. 
The FSC is the largest freely available spoken language understanding dataset that has domain, intent and slots labeling and used a wide range of subjects to record the utterances. The FSC dataset consists of around 30,000 command utterances, each of which  associated with a  3-dimensional vector representing domain, intent and slot.  In total, there are 248 different distinct phrases in the FSC dataset and 5 distinct domains.  The data are split into 23,132 training samples from 77 speakers, 3,118 eval samples from 10 speakers and 3,793 test samples from 10 speakers.

\begin{table}[t]
  \caption{Domain, intent, and slot accuracy predicted by  the hierarchical transformer-based SLU}
  \label{tab:word_styles}
 \centering
\rowcolors{3}{green!25}{yellow!50}
\begin{tabular}{ *5l }    \toprule
\emph{Domain} & \emph{Intent} & Slots&  \\\midrule
98.1    & 99.6  & 99.6  \\\bottomrule
 \hline
\end{tabular}
\end{table}

\begin{table*}[t]
  \caption{The accuracy and number of model parameters for the Transformer,   RNN+SincNet \cite{lugosch2019speech}, and RNN \cite{serdyuk2018towards} models}
   \label{tab:results}
 \centering
\rowcolors{3}{green!25}{yellow!50}
\begin{tabular}{ *5l }    \toprule
\emph{Model} & \emph{Accuracy (\%)} & \# of Parameters&  \\\midrule
Transformer (classification-based, no pre-trained ASR)  & 97.6 & 1,545,987\\ 
Transformer (hierarchical, no pre-trained ASR) & 97.5 & 2,192,320\\ 
RNN+SincNet \cite{lugosch2019speech} (classification-based,  pre-trained ASR, fine-tuned on FSC)  & 97.2 & 2,883,410 \\
RNN+SincNet \cite{lugosch2019speech} (classification-based, no pre-trained ASR)  & 96.1 & 2,883,410 \\
RNN \cite{serdyuk2018towards}(classification-based, no pre-trained ASR)  & 95.3 & 2,534,230 \\\bottomrule
 \hline
\end{tabular}
\end{table*}

 We set up our experiments both for classification-based and hierarchical E2E SLU scenarios.  In classification-based approach, each utterance is assigned one of the 248 distinct labels---each label represents a unique domain-intent-slots phrase. In this approach, we deal with a classification problem and the softmax in the last layer of the decoder outputs the posterior probabilities of labels for each input utterance.  In contrast, in the hierarchical scenario the domain, intent and slots for each utterance are considered as a sequence where we first predict the domain; next the predicted domain is used to predict the intent and the predicted intent is used to predict the slots.  The hierarchical scenario is very similar to the way ASR seq2seq neural architectures work with the difference that instead of using tokens such phonemes, chars, or words, we use sequences of domain, intent, and slots.

\subsection{Results}
We evaluated our E2E SLU neural transformer model using the FSC dataset and compared it with the state-of-the-art  E2E SLU neural model proposed in \cite{lugosch2019speech}, hereafter we call it RNN+SincNet SLU model and its predecessor \cite{serdyuk2018towards}.  We chose \cite{lugosch2019speech} because this model is tested on the FSC dataset,  uses both recurrence and convolutional  architectures that can be compared with the transformer, and  the code source that implements this model is  available which  makes comparison fair.  The  RNN+SincNet SLU model proposed in   \cite{lugosch2019speech}  adapted  from \cite{serdyuk2018towards} with two main differences that makes it more powerful. First, it is  ASR-awareness meaning that it performs phoneme and word recognition before domain, slot, and intent prediction. Second, it deploys  the SincNet layer \cite{Ravanelli2018} that processes the raw audio files  and obviates the need for pre-processing. Moreover, the RNN+SincNet SLU model has the flexibility to be used with or without ASR-awareness and can be  pre-trained for the ASR part on large dataset and used for SLU on small datasets.  The RNN+SincNet model applies five layers of biGRU, three layers of CNN and one layer of linear transformation with dropout and downsampling of factor two in each layer to reduce the time resolution. The RNN+SincNet SLU is, however, a classification-based model which requires the distinct class label for each utterance.

\begin{figure}[h]
  \centering
  \includegraphics[width=.7\linewidth]{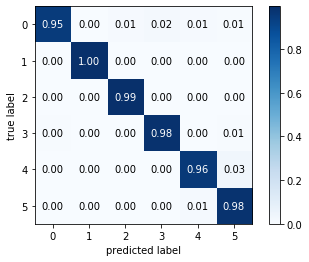}
  \caption{ The confusion matrix of the five domains predicted by the hierarchical SLU transformer model.}
  \label{fig:confusionmatrix}
\end{figure}

We first set our transformer based SLU model to the hierarchical mode and measured the accuracy of domain, intent, and slots. Table~\ref{tab:word_styles} illustrates the accuracy for domain, intent and slots when we use  our model in a hierarchical mode as explained earlier.  In order to evaluate the false negative rate, we also plot the confusion matrix for five distinct domains  as shown in  Figure~\ref{fig:confusionmatrix}. We found  that the most false negative errors stem from domains with very similar audio features like ``increase" and ``decrease" or ``activate" and ``deactivate" but with opposite semantics . This observation underscores that incorporating an ASR-awareness module may reduce the false negative rate where the utterances contain very similar audio features but with different meaning.

Next, we compared our transformer model with RNN based  SLU models   \cite{lugosch2019speech}  and  \cite{serdyuk2018towards}.  We trained these models on the FSC dataset. Table~\ref{tab:results} demonstrates accuracy and the number of parameters for each model.  As shown in Table~\ref{tab:results}, the transformer model outperforms the RNN+SincNet architecture by 1.5\% when both are in the non-pretrained  non-ASR-awareness. We also observed that classification-based  transformer model is marginally better than the hierarchical one. It is not however surprising because the number of classes in the FSC dataset is rather small. Other significant improvement we achieved is the reduction in model size.  The number of parameters in classification-based and hierarchical transformers are 25\% and 46\% smaller than number of parameters in RNN+SincNet models, respectively.  We also compared our model with the RNN model in which  the first two and last two RNN layers are pretrained for phoneme  and word recognition, respectively using Librispeech data and fine-tuned on FSC data; the results show that both  hierarchical and classification transformer based SLU model outperform the pretrained RNN model.  Overall, these results confirm  that the transformers are very competitive compared with recurrence and convolutional-based architectures  because they look at the entire utterance and measure the correlation between different audio units to better extract the semantic information.

\section{Conclusions and future works}
In this paper, we introduced a neural transformer based approach for spoken language understanding.  To the best of our knowledge this is the first time transformers are applied to end-to-end neural SLU.
 We showed that transformers are very competitive neural architectures for end-to-end neural spoken language understanding compared to recurrent and convolutional neural networks.  We evaluated our model on a publicly available dataset and showed the model achieves higher accuracy and reduction in size compared to two competitive models. Because our model is a hierarchical model it can be applied to  variable length domain, intent and slots vectors.  We observed that if the number of unique domain-intent-slot classes is small, the classification-based SLU model delivers better marginal accuracy compared to the hierarchical model. In addition, when we analyzed the false negative rate for domain miss-classification, we found the model has difficulty to predict domains that have very similar audio patterns but very different semantics. Hence, as it was previously observed by other groups, our model may benefit from incorporating an intermediate ASR-awareness module into the neural architecture. One of the unique advantages of transformers is the attention mechanism which is particularly important for SLU.  Transformers compute the correlation between all input vector pairs and thus model knows where to attend to infer semantic information embeded in the audio signal.
 
For future works, we plan: 1- To apply our model to much larger data sets, in particular we will make the use of in-house Alexa data.  2- Adding a multi-task learner by augmenting another decoder to predict intermediate tokens like word-pieces. 3- Using a pre-trained transformer-based ASR model and fine-tune it for the SLU task  
\bibliographystyle{IEEEtran}

\bibliography{mybib}


\end{document}